# Human Gait State Prediction Using Cellular Automata and Classification Using ELM

Vijay Bhaskar Semwal[1,1], Neha Gaud[2] & G.C.Nandi[3]

[1] Department of CSE, Indian Insitute of Inforamtion technology Dharwad, Karnataka, India
{ vsemwal@iiitdwd.ac.in }

[2] Institute Of Computer Science, Vikram University Ujjain , M.P., India

[3] Robotics & AI Lab, Indian Insitute of Inforamtion Technology Allahabad , U.P., India
{ gcnandi@iiita.ac.in }

**Abstract.** In this research article, we have reported periodic cellular automata rules for different gait state prediction and classification of the gait data using extreme machine Leaning (ELM). This research is the first attempt to use cellular automaton to understand the complexity of bipedal walk. Due to non-linearity, varying configurations throughout the gait cycle and the passive joint located at the unilateral foot-ground contact in bipedal walk resulting variation of dynamic descriptions and control laws from phase to phase for human gait is making difficult to predict the bipedal walk states. We have designed the cellular automata rules which will predict the next gait state of bipedal steps based on the previous two neighbour states. We have designed cellular automata rules for normal walk. The state prediction will help to correctly design the bipedal walk. The normal walk depends on next two states and has total 8 states. We have considered the current and previous states to predict next state. So we have formulated 16 rules using cellular automata, 8 rules for each leg. The priority order maintained using the fact that if right leg in swing phase then left leg will be in stance phase. To validate the model we have classified the gait Data using ELM [1] and achieved accuracy 60%. We have explored the trajectories and compares with another gait trajectories. Finally we have presented the error analysis for different joints.

## 1. Introduction

The bipedal walk is the combination of 8 different sub phases during normal walk. In running, we observed that it is a combination of 4 different sub phases [2]. The Human walk is inherently unstable and non linear due to high degrees of non-linearity, high dimensionality, under actuation (in swing phase), over actuation (in stance phase). The modern robotics industries have given boost the development of such robot which can walk in unconstrained environment similar to human. The modern robots are not capable enough to walk effectively in unconstraint environment. The next state predication for robotic walk is very difficult. The bipedal robot generally used to suffer from the singularity configuration. Using pseudo inverse, we used to deal such configuration [3].

The equation of linear system can be represented in following form:

$$Ax = y - (1)$$

$$A \text{ is singluar if } \|A\| = 0 - (2)$$

$$Pseudo\ Inverse\ x = (A^tA)^{-1}A^ty - (3)$$

This research is attempts to predict the state of robot walking. The walking pattern is very unique to each human being [4]. The human acquired this behavior through learning. The gait study is widely using in Biometric identification [3] [5], artificial limbs generation [6], Robotics walk [7] [8] and development of modern data driven computational walking model which can walk similar to human [9].

We have presented the cellular automata model as generalized predictive model. The model will be able to predict the next state based on current state and previous state. It will help the robot to plan the next state. Total 8! Permutation is possible. The model is able to predict the state on any terrain. In this case we have considered only states in terms of joints angle value we would not referred the terrain. Cellular automata are mathematical tool [10] for modeling a system that evolves with certain set of rule. It merges the specifications for discrete switching logic and continuous dynamics behaviors of any dynamical system. So, it is an appropriate model for predicting a human bipedal trajectory in terms of CA because a bipedal locomotion trajectory is also a combination of continuous and discrete phases [11] and a stable walk can be obtained using CA mathematical model. This bipedal trajectory very precisely can be designed using cellular automata.

The paper is organized into 4 parts as following. The next section is methodology section which is description of cellular automata rule and ELM algorithm description. The third section is results section. The fourth section is verification of results and final section is conclusion, discussion and future scope.

## 2. Methodology

To modelling the system we have used a bottom-up approach. We have first constructed the atomic component and modelled their behaviour using Cellular automaton. We have merged all atomic component together and converted into composite components. The interactions can be formulated between composite and atomic components using the well-defined semantics of algebra of connectors and mainly the causal chain type of interaction. Later, we assigned the priorities which impose the certain ordering on the type of interactions. The priority also can help to avoid the deadlock. Before, we develop the cellular automata and their design, it is important to understand the different sub phases of one complete gait. Mostly the gait data is used in the medical and health care sector. The gait cycle usually takes 1-1.2 s to complete. The gait is a time series data and broadly it has two phases Stance and Swing. The gait further can be dived into eight sub phases [12] [13]. As gait is time series data, so each sub phases takes certain percentage of entire gait cycle. The swing phases generally takes 40% of whole gait cycle and stance phase takes 60% of complete gait cycle. The percentage wise division of human gait is divided into following sub phases [14] [15] [16]:

Stance phase:
1. Initial Contact– IC[0-2%]
2. Loading Response– LR[2-10%]
3. Mid Stance– MS[10-30%]
4. Terminal Stance– TS[30-50%]

Swing phase:
5. Pre Swing– PS[50-60%]
6. Initial Swing– IS[60-73%]
7. Mid Swing - MS[73-87%]
8. Terminal Swing – TS[87-100%]

The initial contact sub phase accounts for a very small percentage of the complete gait cycle, hence it is merged with loading response phase without loss of generality and we called the new merged state as initial contact. Similarly, we have merged the pre-swing and initial swing in one combined state named it as initial swing [17].

### 2.1 Design of Atomic Components for CA

The left and right legs can be decomposed into three atomic components named hip, knee and ankle. So, we have total six atomic components i.e. each leg should have three atomic component. For two legs we will have two composite components. To express the nature of each atomic component is given by a six states (3 for each swing and stance states) in cellular automaton. In our cellular automata model the states starts from stance phase which lead for automatically swing phase for right leg [18]. All shifting from one state to other happens in synchronisation. So, the phase order during leg moment will be the left leg goes into swing phase and right will go in stance phase. [19].

### 2.2 Developing Cellular Automata(CA)

Here we have written 16 CA rules to determine the state of atomic components of one leg with the help of second leg. It will be among one of the eight states so there will be a total of 16 rules. 1000 can be seen as two parts 1+000(leg + Sub phase) which means right leg is in initial contact.

We have assumed binary state of movement of atomic components of a leg (Ankle, Knee, Hip) is either in motion or in rest. So we consider binary stage 0 and 1 for each component. 0 represents atomic components are in rest and 1 represents atomic components are in motion [19]. Since, there are three atomic components and each have two state either 0 or 1. So, there will be a total of eight states. During human locomotion each leg passes through eight sub phases [20].

CA is discrete dynamic systems. CA's are said to be discrete because they operate in finite space and time and with properties that can have only a finite number of states. CA's are said to be dynamic because they exhibit dynamic behaviors. Eq. 4 is the representation of state prediction. Where S(t) represents the current state. S is the set of all possible discrete states of our gait model for us it is 8. And Eq. 5 is the prediction of next state from the previous two states.

S: Finte set of state i. e. discreter variable

$$S = \{IC, LR, MS, TST, PSW, ISW, MSW, TSW\} \quad - (4)$$

$$S(t + 1) = \{S(t), S(t - 1)\} \quad - (5)$$

Consider the state S={1,2,3,4,5,6,7,8} so we have assumed 8 discrete states here. In this work we have considered 8 discrete state as 8 neighbors. Eq. 5 is the state prediction in case of Normal Walk and brisk walk. Each leg which is considered as atomic component passes through the 3 discrete states named initial contact, mid stance, terminal stance in synchronization. The phase transfer between the left and right leg happens alternatively. The swing phase gets complete once the left leg returns back to the stance phase [21]. The atomic components of the left leg are shown here ankle, knee and hip. We have assumed binary state of movement of atomic components of a leg (Ankle, Knee, Hip) is either in motion or in rest. So we consider binary stage 0 and 1 for each component. 0 represents atomic components are in rest and 1 represents atomic components are in motion. Since, there are three atomic components and each have two state either 0 or 1. So, there will be a total of eight states (Refer Table 1).

### 2.3 CA Rules

Here we have written 16 CA (equation-6 to equation 14) rules to determine the state of atomic components of one leg with the help of second leg. All the states are represented using 4-bit stream. First bit represent the leg that if the fourth bit is zero it represents left leg whereas if the fourth bit is 1 it represents the right leg. Other three bits represents the sub phases of that leg. It will be among one of the eight states so there will be a total of 16 rules. 1000 can be seen as two parts 1+000(leg + Sub phase) which means right leg is in initial contact. The neighbor row represents the state or phase of another leg whereas the rule row depicts the state of atomic components of that leg. Set of rules to determine the state of locomotion. Cellular automata rule Rule-8, universal, generalizes Rule for left and right leg during normal walk. Following are the states relation between left and right leg.

Left_Leg_Stance->Right_Leg_Swing – (6)
Left_Leg_Swing->Right_Leg_Stance – (7)
Left_Leg_IC ->Right_Leg_PSw – (8)
Left_Leg_MS->Right_Leg_Msw – (9)
Left_Leg_TS->Right_Leg_TSw– (10)
Left_Leg_PSw->Right_Leg_LR– (11)
Left_Leg_ISw->Right_Leg_MS– (12)
Left_Leg_MSw->Right_Leg_TS– (13)
Left_Leg_TSw->Right_Leg_IC– (14)

Tables 1 to 3 are the binary state representation of bipedal gait and it is the novel contribution. It will help to predict the next state of robot. Fig.1 is the transaction of leg state using cellular automata. The unique approach to model the human gait presented here. It is able to model the normal human gait within a negotiable degree of error. Here we have written 16 CA rules to determine the state of atomic components of one leg with the help of second leg. All the states are represented using 4-bit stream. First bit represent the leg that if the fourth bit is zero it represents left leg whereas if the fourth bit is 1 it represents the right leg. Other three bits represents the sub phases of that leg. It will be among one of the eight states so there will be a total of 16 rules. 1000 can be seen as two parts 1+000(leg + Sub phase) which means right leg is in initial contact. The neighbor row represents the state or phase of another leg whereas the rule row depicts the state of atomic components of that leg.

Table.1. Binary State Representation of Bipedal Gait 8 states

| Number | 7 | 6 | 5 | 4 | 3 | 2 | 1 | 0 |
|---|---|---|---|---|---|---|---|---|
| Neighborhood | 111 | 110 | 101 | 100 | 011 | 010 | 001 | 000 |
| Rule Result | TS | MS | IS | PS | TS | MS | LR | IC |

Table.2. Cellular Automata state prediction for Left leg

| Number | 7 | 6 | 5 | 4 | 3 | 2 | 1 | 0 |
|---|---|---|---|---|---|---|---|---|
| Neighbor | 0111 | 0110 | 0101 | 0100 | 0011 | 0010 | 0001 | 0000 |
| Rule Result | 1011 | 1010 | 1001 | 1000 | 1111 | 1110 | 1101 | 1100 |

Table.3. Cellular Automata state prediction for Left leg

| Number | 15 | 14 | 13 | 12 | 11 | 10 | 9 | 8 |
|---|---|---|---|---|---|---|---|---|
| Neighbor | 1011 | 1010 | 1001 | 1000 | 1111 | 1110 | 1101 | 1100 |
| Rule Result | 0111 | 0110 | 0101 | 0100 | 0011 | 0010 | 0001 | 0000 |

CA are discrete dynamic systems.
- CA's are said to be discrete because they operate in finite space and time and with properties that can have only a finite number of states.
- CA's are said to be dynamic because they exhibit dynamic behaviors.

**2.4. Basic ELM Classifier:** Hung, et al 2004, 2006 [1] [3] has proposed the ELM, which is fast and we called learning without iteration tuning. For given non-constant piece wise continuous function g, if continuous target function f(x) can be approximated by SLFNS with some adjustable weights

Given a training set $\{(a_i, b_i) \mid a_i \in R^d, b_i \in R^m, i=1,..,N\}$, hidden node output function G(a, b, x), and the number of hidden nodes L,

Step-1: Randomly hidden node parameter $(a_i, b_i)$, i=1,…,N.
Step 2- Output of hidden layer $H=[h(x_1), h(x_2)… h(x_N)]^T$
Step 3-The out weight β.

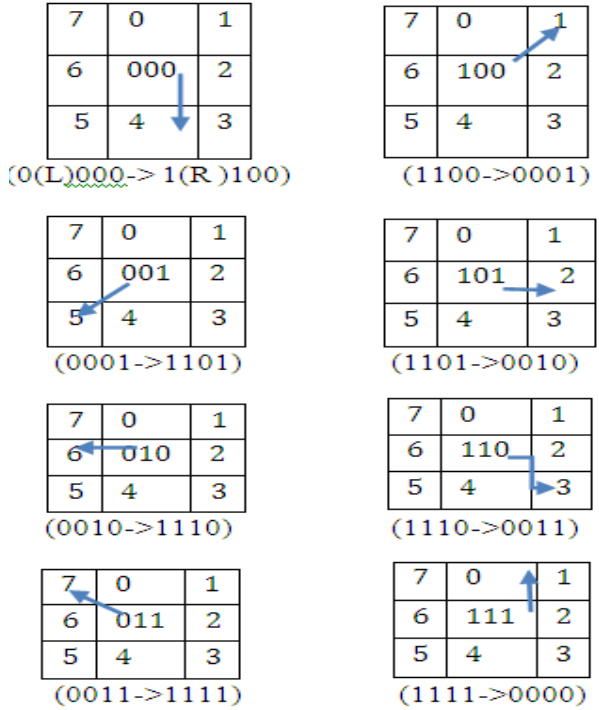

Fig.1.　　Transaction of leg state using CA

## 3. Experiments Results and Verification:

In this paper, we will be looking on the results and outcome of the work done using cellular automata for the state prediction of bipedal walk. For this we have divided the human gait cycle data into different sub phases for different joint angle right and left ankle, knee and hip joints. These are the equations for all the six joints that are left hip, left knee, left ankle and right hip, right knee and right ankle corresponding to each sub-phase of the gait cycle [22][23].

Fig. 2 is classification accuracy of cellular automata based gait using ELM, SVM and linear Regression. ELM based classification has our performed with 65% to other classifiers. Similarly it has proved the cellular automata based Gait state predication is giving a better walking pattern. Fig.3 is the stick diagram to validate the cellular automata rules. In this model we have applied the joint trajectories. We have taken the input thigh and stride length. It is showing the state predicted through cellular automata is correct. Fig.4 is the different gait states prediction of Gait2354 model of Opensim.  Fig.5 shows the hip joints limit cycle to validate the stability of state generated trajectory of cellular automata. The limit cycle is used to measure the stability of model. The trajectories generated using the model for left and right hip is depicted in Fig.4. It is following the cycle which justify that the hip joints trajectories are stable .Fig. 5 is the comparison of trajectories of our model with other well established trajectories and our previous work generated trajectories [24][25]. The generated trajectories are very close to other established joints trajectories.

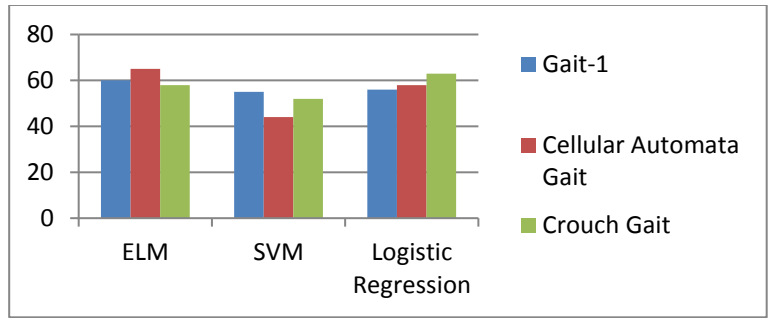

Fig.2. Classification Result of Different gait using ELM, SVM, and Linear Regression.

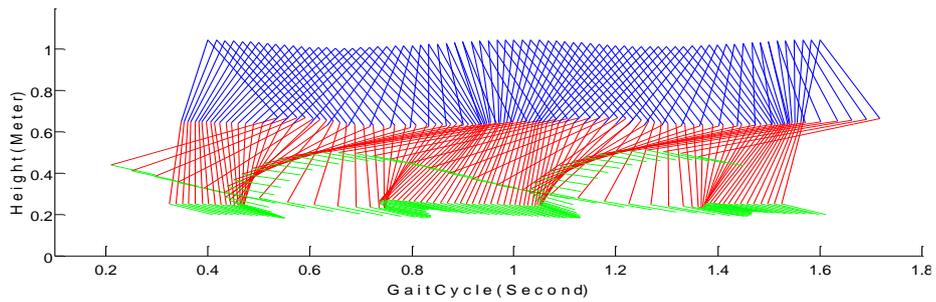

Fig.3. Stick diagram for both leg gait pattern of cellular automata model

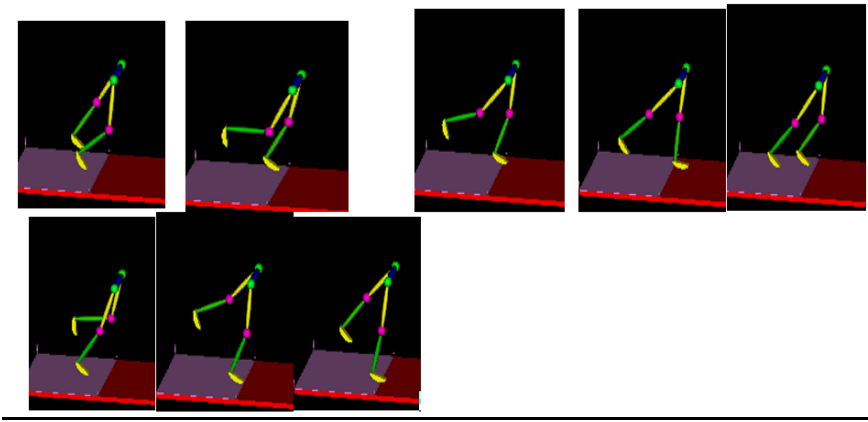

Fig.4. The transaction and State Prediction of leg state using cellular automata

## 4. Error Analysis:

In this section we have validated the trajectories generated through cellular automata. To validate the correctness of model, it is important to perform the error analysis of predicated state using cellular automata model. The analytical human data base model and (stable for 4 degree) and the Cellular Automata Model (stable for 4 degree). We are now going to calculate the error in Analytical model with respect to Human data. We have then plotted this output in figure 6. From the figure, it is clearly visible that the mean percentage error has the maximum correlation for both knee and hip. Hence, this error is suitable for training the neural network for finding the error pattern recognition [26].

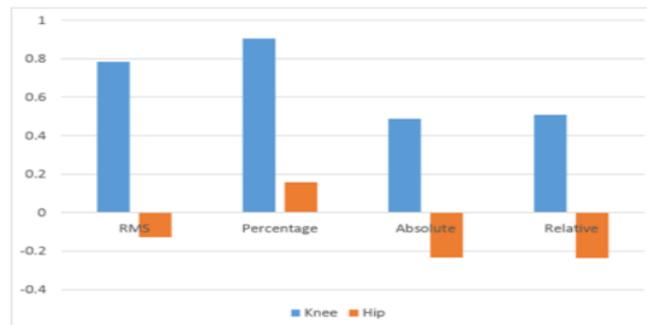

Fig.5. Correlation between various error with Spearmen's error

In this section we have compared the limit cycle behaviour between Hip and Knee joints for the cellular automata generated trajectories. We have identified that the base model is unstable below some degree. For comparison, we have used walking along some degrees of inclination. As mentioned in the earlier section, the limit cycle behaviour clearly shows that the system is unstable at some degrees of inclination. Now as seen from the fig.6 the limit cycle behaviour shows that the same system is stable at some degrees angle of inclination [27]. This comparison clearly shows while the base model fails to justify stable walking by a real subject under certain conditions, the cellular model fully supports the same.

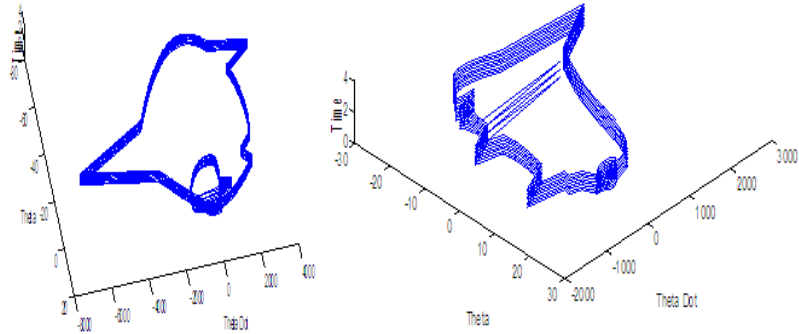

Fig.6.  (a) Cellular Automata Model Knee LC  (b) Cellular Automata Model Hip LC

## 5. Conclusion & Future Discussion

We have presented the unique approach to model the human gait using cellular automata. The proposed model is able to design the human gait with some negotiable degree of error. Here we have written 16 CA rules to determine the state of atomic components of one leg with the help of second leg. All the states are represented using 4-bit stream. First bit represent the respective leg. If the fourth bit is zero it represents left leg whereas if the fourth bit is 1 it represents the right leg. Other three bits represents the sub phases of that leg. It will be among one of the eight states so there will be a total of 16 rules. 1000 can be seen as two parts 1+000(leg + Sub phase) which means right leg is in initial contact. The Neighbour row represents the state or phase of another leg whereas the rule row depicts the state of atomic components of that leg. The cellular automata based model help to predict the next state of robot walking. It will generate the trajectories which we have verified using limit cycle curve and comparison with other trajectories. The covariance matrix is calculated about the parameter to prove how the cellular automata generated trajectories is close to other normal walk trajectories. We have achieved classification accuracy 60% through ELM for cellular automata based walk. The ELM based classifier it bext among all classifier and very fast.

**Future Discussion:** The cellular automata can be utilized for many other learning behaviors like push recover [27]. The behavior can be explored far better and can be implemented on bipedal robot model. The state prediction is unconstrained environment is very tough. It requires the human intelligence and neuron muscular co-ordination.

# References

1. Huang, Guang-Bin, Qin-Yu Zhu, and Chee-Kheong Siew. "Extreme learning machine: a new learning scheme of feedforward neural networks." Neural Networks, 2004. Proceedings. 2004 IEEE International Joint Conference on. Vol. 2. IEEE, 2004.
2. Semwal, Vijay Bhaskar, et al. "Design of Vector Field for Different Subphases of Gait and Regeneration of Gait Pattern." IEEE Transactions on Automation Science and Engineering, vol. PP, no. 99, pp. 1-7, 2016.
3. Huang, Guang-Bin, Lei Chen, and Chee Kheong Siew. "Universal approximation using incremental constructive feedforward networks with random hidden nodes." IEEE Trans. Neural Networks 17.4 (2006): 879-892.
4. Raj, Manish, Vijay BhaskarSemwal and G. C. Nandi. "Bidirectional association of joint angle trajectories for humanoid locomotion: the restricted Boltzmann machine approach," Neural Computing and Applications, pp. 1-9, 2016.
5. Semwal, Vijay Bhaskar, Manish Raj, and G. C. Nandi. "Biometric gait identification based on a multilayer perceptron." Robotics and Autonomous Systems, vol. 65, pp. 65-75, 2015.
6. S. C. Mukhopadhyay, "Wearable Sensors for Human Activity Monitoring: A Review," in IEEE Sensors Journal, vol. 15, no. 3, pp. 1321-1330, March 2015.
7. Zhang, Zhaoxiang, Maodi Hu, and Yunhong Wang. "A survey of advances in biometric gait recognition." In Biometric Recognition, pp. 150-158. Springer Berlin Heidelberg, 2011.
8. Gupta, Jay Prakash, et al. "Human activity recognition using gait pattern." International Journal of Computer Vision and Image Processing (IJCVIP) 3.3 (2013): 31-53.
9. Vijay Bhaskar Semwal "Data Driven Computational Model for Bipedal Walking and Push Recovery". DOI: 10.13140/RG.2.2.18651.26403.
10. M. Raj, V. Bhaskar Semwal, G.C.Nandi. Hybrid Model for Passive Locomotion Control of a Biped Humanoid:The Artificial Neural Network Approach, International Journal of Interactive Multimedia and Artificial Intelligence, (2017).
11. Semwal, Vijay Bhaskar, et al. "Biologically-inspired push recovery capable bipedal locomotion modeling through hybrid automata," Robotics and Autonomous Systems, vol. 70, pp.181-190, 2015.
12. Huang, Guang-Bin, et al. "Extreme learning machine for regression and multiclass classification." IEEE Transactions on Systems, Man, and Cybernetics, Part B (Cybernetics) 42.2 (2012): 513-529.
13. Wang, Chen, Junping Zhang, Liang Wang, Jian Pu, and Xiaoru Yuan. "Human identification using temporal information preserving gait template." Pattern Analysis and Machine Intelligence, IEEE Transactions on 34, no. 11 (2012): 2164-2176.
14. Wang, Liang, Tieniu Tan, HuazhongNing, and Weiming Hu. "Silhouette analysis-based gait recognition for human identification." Pattern Analysis and Machine Intelligence, IEEE Transactions on 25, no. 12 (2003): 1505-1518.
15. Semwal, Vijay Bhaskar, et al. "An optimized feature selection technique based on incremental feature analysis for bio-metric gait data classification." Multimedia Tools and Applications (2016): 1-19.
16. A. Nag, S. C. Mukhopadhyay and J. Kosel, "Wearable Flexible Sensors: A Review," in IEEE Sensors Journal, vol. 17, no. 13, pp. 3949-3960, July1, 1 2017.
17. Sinnet, R. W, Powell, M. J., Shah, R. P., Ames, A. D. "A human-inspired hybrid control approach to bipedal robotic walking." In 18th IFAC World Congress (pp. 6904-11), 2011.
18. Semwal, V.B., Nandi, G.C., "Toward Developing a Computational Model for Bipedal Push Recovery:A Brief," Sensors Journal, IEEE, vol.15, no.4, pp.2021-2022, 2015.
19. Parashar A, Parashar A, Goyal S. "Push Recovery for Humanoid Robot in Dynamic Environment and Classifying the Data Using K-Mean," International Journal of Interactive Multimedia and Artificial Intelligence, vol. 4, no. 2, pp. 29-34, 2016.

20. Semwal, Vijay Bhaskar, and Gora Chand Nandi. "Generation of Joint Trajectories Using Hybrid Automate-Based Model: A Rocking Block-Based Approach,"IEEE Sensors Journal, vol. 16, no. 14, pp. 5805-5816, 2016.
21. Semwal, V.B.; Katiyar, S.A.; Chakraborty, P.; Nandi, G.C., "Biped model based on human Gait pattern parameters for sagittal plane movement," Control, Automation, Robotics and Embedded Systems (CARE), 2013 International Conference on , vol., no., pp.1,5, 16-18 Dec. 2013.
22. Raj, Manish, Vijay BhaskarSemwal and G. C. Nandi. "Multiobjective optimized bipedal locomotion," International Journal of Machine Learning and Cybernetics, pp. 1-17, 2017.
23. Nandi, Gora Chand, et al. "Modeling bipedal locomotion trajectories using hybrid automata," Region 10 Conference (TENCON), 2016 IEEE. IEEE, 2016.
24. Vijay BhaskarSemwal, Pavan Chakraborty and G.C. Nandi., "Biped model based on human Gait pattern parameters for sagittal plane movement," IEEE International Conference on Control, Automation, Robotics and Embedded Systems (CARE), 2013, pp. 1-5.
25. Vijay BhaskarSemwal and G.C. Nandi, "Robust and more accurate feature and classification using deep neural network," Neural Computing and Application, vol. 28, no. 3, pp. 565-574.
26. Vijay BhaskarSemwal and G.C. Nandi., "Study of humanoid Push recovery based on experiments," IEEE International Conference on Control, Automation, Robotics and Embedded Systems (CARE), 2013.
27. Vijay BhaskarSemwal, Pavan Chakraborty and G.C. Nandi, "Less computationally intensive fuzzy logic (type-1)-based controller for humanoid push recovery," Robotics and Autonomous Systems, vol. 63, Part 1, pp. 122-135, 2015.